\documentclass[pdflatex,sn-basic]{sn-jnl}

\usepackage{graphicx}%
\usepackage{multirow}%
\usepackage{amsmath,amssymb,amsfonts}%
\usepackage{amsthm}%
\usepackage{mathrsfs}%
\usepackage[title]{appendix}%
\usepackage{xcolor}%
\usepackage{textcomp}%
\usepackage{manyfoot}%
\usepackage{booktabs}%
\usepackage{algorithm}%
\usepackage{algorithmicx}%
\usepackage{algpseudocode}%
\usepackage{listings}%
\usepackage{amsmath,amssymb,amsfonts}
\usepackage{booktabs}
\usepackage{multirow}
\usepackage{array}

\theoremstyle{thmstyleone}%
\theoremstyle{thmstyletwo}%

\theoremstyle{thmstylethree}%

\begin{document}

\title[Article Title]{Multi-Source Multi-View Graph Domain Adaptation with Hyperbolic Residual Encoding for Cross-Site MDD Identification from rs-fMRI}


\author[1,3]{\fnm{Zhanpeng} \sur{Zheng}}
\equalcont{These authors contributed equally to this work.}

\author*[1,2,3]{\fnm{Xiran} \sur{Chen}}\email{lightphenexx@outlook.com}
\equalcont{These authors contributed equally to this work.}

\author[4]{\fnm{Haiteng} \sur{Jiang}}

\author[5]{\fnm{Renjie} \sur{Tian}}

\author[6]{\fnm{Qinyu} \sur{Cai}}

\author[7]{\fnm{Jiexi} \sur{Liu}}

\author[3]{\fnm{Xiaofeng} \sur{Chen}}

\author*[1]{\fnm{Weikai} \sur{Li}}\email{leeweikai@outlook.com}

\author*[2]{\fnm{Yansu} \sur{Wang}}\email{wangyansu@uestc.edu.cn}

\affil[1]{\orgdiv{School of Computer and Artificial Intelligence, Shandong Jianzhu University, Shandong, China}}

\affil[2]{\orgdiv{Institute of Fundamental and Frontier Sciences, University of Electronic Science and Technology of China, Chengdu, Sichuan, China}}

\affil[3]{\orgdiv{School of Mathematics and Statistics, Chongqing Jiaotong University, Chongqing, China}}

\affil[4]{\orgdiv{State Key Laboratory of Brain-machine Intelligence, Zhejiang University, Hangzhou, China}}

\affil[5]{\orgdiv{Institute of Computer Vision and Traffic Image Understanding, School of Information Science and Engineering, Chongqing Jiaotong University, Chongqing, China}}

\affil[6]{\orgdiv{School of Life Sciences, Westlake University, Hangzhou, China}}

\affil[7]{\orgdiv{School of Computer and Artificial Intelligence,
Nanjing University of Finance and Economics, Nanjing, China}}

\abstract{Cross-site identification of major depressive disorder (MDD) from resting-state functional magnetic resonance imaging (rs-fMRI) is hindered by inter-site distribution shifts and heterogeneous functional connectivity (FC) views. These views capture complementary neural relationships but exhibit distinct site biases and graph topologies, complicating alignment without sacrificing disease-relevant information or cross-view consistency. Existing studies largely treat multi-view connectome learning and cross-site adaptation separately. To the best of our knowledge, few studies have jointly modeled multiple FC views under multi-source unsupervised domain adaptation for cross-site rs-fMRI-based MDD classification. We construct Pearson correlation, sparse representation, and Granger causality graphs, each encoded by a view-specific graph attention network. Dual-stream adaptive fusion explicitly integrates pairwise cross-view interactions, followed by lightweight hyperbolic residual encoding for curvature-aware representation refinement. Class-wise Cauchy--Schwarz alignment reduces inter-source and source-target discrepancies, complemented by adversarial learning, information maximization, and confidence-aware pseudo-labeling. Across seven unlabeled target domains, our framework achieves 73.60\% mean accuracy and 71.90\% AUC, demonstrating effective generalization under heterogeneous acquisition conditions. These results highlight the effectiveness of unified heterogeneous-view modeling, curvature-aware refinement, and multi-source domain adaptation for cross-site MDD identification. The source code is at https://github.com/OPUS-Lightphenexx/MM-HyperGDA}




\maketitle

\section{Introduction}

Major depressive disorder (MDD) is among the most prevalent and
debilitating mental health conditions worldwide \cite{marx2023major},
yet its diagnosis still relies largely on clinical interviews and
symptom-based assessment. Resting-state functional magnetic resonance
imaging (rs-fMRI) offers a promising means of identifying objective
neural biomarkers through intrinsic functional connectivity. However,
reliable rs-fMRI-based MDD diagnosis remains challenging.

The practical application of rs-fMRI-based MDD classification is hindered by three increasingly intractable intertwined obstacles. First, data collected from different acquisition sites exhibit severe inter-site distribution shifts, which drastically degrade the generalizability of trained models\cite{dansereau2017statistical}. Second, the characterization of brain functional connectivity is inherently multi-faceted; different construction methods \cite{bastos2016tutorial} capture complementary neural relationships, ranging from undirected synchrony to directed causal influences. Critically, these diverse views exhibit distinct site-specific biases and graph topologies \cite{friston2011functional}. This heterogeneity makes multi-source domain alignment extremely difficult, as naive adaptation strategies risk either erasing disease-relevant, view-specific information or disrupting the essential cross-view consistency among complementary representations. Third, hierarchically organized brain networks may incur representation distortion when modeled exclusively in conventional Euclidean spaces.

While unsupervised domain adaptation can mitigate source--target
distribution shifts, extending it simultaneously to multiple sources and
multiple functional-connectivity views is substantially more challenging.
First, labeled source sites may differ in acquisition protocols, cohort
characteristics, and class-conditional distributions; directly pooling
them can obscure source-specific structure and induce negative transfer.
Second, Pearson correlation, sparse representation, and Granger causality
produce heterogeneous feature distributions and graph topologies.
Forcing these views into a common alignment space may suppress
disease-relevant view-specific information, whereas adapting them
independently may overlook their complementary cross-view dependencies.
The joint setting therefore requires coordinating inter-source alignment,
source--target adaptation, and cross-view integration while preserving
both class-discriminative structure and view diversity.

To address these coupled challenges, we formulate cross-site MDD
identification as a unified multi-source, multi-view graph domain
adaptation problem. View-specific graph attention encoders preserve the
distinct topology of each connectivity construction, while pairwise
cross-view interaction and dual-stream adaptive fusion integrate
complementary information without collapsing view-specific
representations. Class-wise Cauchy--Schwarz alignment jointly reduces
inter-source and source--target discrepancies, while adversarial learning,
information maximization, and confidence-aware pseudo-labeling provide
complementary global and target-structure adaptation signals. Finally, a
lightweight post-fusion hyperbolic residual encoder performs
curvature-aware refinement of the fused subject-level representation.

In summary, the main contributions of this work are summarized as follows:
\begin{enumerate}
    \item We formulate cross-site MDD identification as a multi-source,
multi-view unsupervised graph domain adaptation problem, where
inter-source heterogeneity, source--target shifts, and view-dependent
distribution biases are jointly addressed without collapsing
class-discriminative structure or view diversity.

    \item We develop a unified architecture that preserves view-specific
    graph structures, explicitly models pairwise cross-view dependencies,
    and adaptively integrates direct and interaction-based representations
    before post-fusion curvature-aware refinement.

    \item We integrate class-conditional source alignment,
    source-target distribution matching, domain-adversarial
    learning, information maximization, and confidence-aware
    pseudo-labeling into a unified two-stage optimization framework,
    and evaluate the framework across seven target sites.
\end{enumerate}

\section{Related Work}

\subsection{Cross-Site Domain-Adaptive Diagnosis}

Cross-site rs-fMRI analysis is affected by scanner, protocol, and cohort
differences that induce substantial domain shifts
\cite{dansereau2017statistical}. Unsupervised domain adaptation uses
labeled source data and unlabeled target data to learn transferable
representations. UFA-Net \cite{fang2023unsupervised} combines
attention-guided graph learning with distribution alignment for
cross-site MDD identification, whereas AUFA \cite{ma2026aufa}
introduces augmentation-based self-optimization to reduce source-model
overfitting. In the broader context of cross-site neuropsychiatric
diagnosis, H2MSDA \cite{luo2025multi} addresses multi-source adaptation
for ASD through hierarchical alignment across labeled sources and an
unlabeled target. However, these methods primarily rely on a single
functional-connectivity construction and therefore do not explicitly
address view-dependent domain shifts or interactions among heterogeneous
connectivity representations. This limitation becomes more pronounced
when source sites exhibit different class-conditional distributions.

\subsection{Multi-View Functional Connectivity Learning}

Different connectivity constructions characterize complementary neural
relationships. Pearson correlation captures undirected synchronization,
sparse representation models reconstruction-based dependence, and
Granger causality describes directed temporal predictability.
MFCP \cite{ruan2026identification} integrates these three views through
multiple graph convolutional branches for MDD identification. Related
multi-connectome methods also demonstrate the value of heterogeneous graph
views in brain-disorder classification
\cite{yang2024towards,zhou2024multipattern}. Nevertheless, these
approaches generally assume matched training and test distributions and
thus do not address how multiple connectivity views should be jointly
adapted when labeled source sites and an unlabeled target exhibit distinct,
view-dependent shifts.

\subsection{Hyperbolic Representation Learning}

Hyperbolic spaces provide compact representations for data with latent
hierarchical organization and have been adopted in neural and graph
representation learning
\cite{nickel2017poincare,ganea2018hyperbolic,chami2019hgcn}.
Recent neuroimaging studies have also explored hyperbolic geometry for
cross-site alignment \cite{luo2025multi}. However, existing methods
typically operate on a single connectivity construction and do not examine
how heterogeneous graph views should be integrated before geometric
encoding. In contrast, our framework preserves view-specific structures,
models pairwise cross-view dependencies, performs adaptive dual-stream
fusion, and applies lightweight Poincar\'e-ball-constrained residual
refinement while jointly reducing inter-source and source--target
discrepancies.

\section{Materials and Data Preprocessing}

\subsection{Dataset and Evaluation Protocol}

Experiments were conducted on the REST-meta-MDD multi-site resting-state
functional magnetic resonance imaging dataset for major depressive
disorder identification \cite{yan2019restmetamdd}. Site20 and Site21 were treated as two labeled
source domains. Site6, Site11, Site12, Site16, Site19, Site22, and
Site25 were independently treated as unlabeled target domains.

For each target-site experiment, all target samples participated in
adaptation without ground-truth labels. Target labels were not used in
source pretraining, adaptation losses, pseudo-label construction, or
parameter optimization, and were used only after training to compute ACC
and AUC. The experiments therefore followed a transductive multi-source
unsupervised domain adaptation setting.

Each target site was adapted and evaluated independently. Accuracy
(ACC) was calculated from the final predicted labels, and the area under
the receiver operating characteristic curve (AUC) was calculated from
the predicted probabilities. Site-wise results were reported for all
seven target domains. Macro-averaged ACC and AUC were further
calculated by assigning equal weight to each target site.

\begin{table}[h]
\centering
\caption{Sample statistics of the source and target sites. HC denotes
healthy controls.}
\label{tab:data_stats}
\begin{tabular}{lccc}
\toprule
Site & HC & MDD & Total\\
\midrule
Site20 & 251 & 282 & 533\\
Site21 & 70 & 86 & 156\\
Site6  & 15 & 15 & 30\\
Site11 & 29 & 32 & 61\\
Site12 & 6  & 32 & 38\\
Site16 & 31 & 31 & 62\\
Site19 & 36 & 51 & 87\\
Site22 & 20 & 30 & 50\\
Site25 & 63 & 89 & 152\\
\bottomrule
\end{tabular}
\end{table}

\subsection{ROI Time Series and Multi-View Network Construction}

We used publicly released preprocessed ROI time series from
REST-meta-MDD. Following the reported DPARSF pipeline, the first ten
volumes were removed, followed by head-motion correction, spatial
normalization and smoothing, nuisance-signal regression, and band-pass
filtering
\cite{yan2010dparsf,power2012motion,hallquist2013nuisance}. Mean BOLD
signals were extracted from 116 AAL regions. We directly loaded these
AAL116 signals, z-standardized each regional series across time, and
performed no ROI re-binning or dimensional aggregation.

Accordingly, each subject was represented by an AAL116 regional
time-series matrix. Let
$\mathbf{X}\in\mathbb{R}^{T\times 116}$ denote the regional time-series
matrix, where $T$ is the number of temporal observations.

To characterize complementary forms of interregional interaction, three
functional connectivity views were constructed for every subject:
Pearson correlation (PC), sparse representation (SR), and Granger
causality mapping (GCM).

\paragraph{Pearson correlation view.}

The PC view characterizes undirected linear synchronization between
regional time series. The connectivity value between regions $i$ and
$j$ is calculated as
\begin{equation}
\mathbf{A}^{\mathrm{PC}}_{ij}
=
\left|
\frac{
\sum_t (x_{ti}-\bar{x}_{i})(x_{tj}-\bar{x}_{j})
}{
\sqrt{\sum_t (x_{ti}-\bar{x}_{i})^2}
\sqrt{\sum_t (x_{tj}-\bar{x}_{j})^2}
}
\right|.
\end{equation}

\paragraph{Sparse representation view.}

The SR view characterizes reconstruction-based dependencies. For each
regional signal $\mathbf{x}_i$, the remaining regional signals are used
to obtain a sparse reconstruction \cite{tibshirani1996lasso}:
\begin{equation}
\hat{\boldsymbol{\beta}}_i
=
\arg\min_{\boldsymbol{\beta}_i}
\left\|
\mathbf{x}_i-\mathbf{X}_{-i}\boldsymbol{\beta}_i
\right\|_2^2
+
\lambda_{\mathrm{SR}}
\left\|
\boldsymbol{\beta}_i
\right\|_1 .
\end{equation}
The resulting coefficients are placed into a connectivity matrix,
symmetrized, and normalized to obtain $\mathbf{A}^{\mathrm{SR}}$. In our
implementation, $\lambda_{\mathrm{SR}}$ was set to 0.01.

\paragraph{Granger causality mapping view.}

The GCM view characterizes directed temporal predictability \cite{granger1969causal,seth2015granger}. For a pair
of regions, the method measures whether the lagged activity of one
region improves prediction of the activity of the other region. The
relative reduction in residual variance is used as the directed
connectivity score, producing $\mathbf{A}^{\mathrm{GCM}}$. In our
implementation, the GCM lag order was set to 1.

Each connectivity matrix was converted into a weighted sparse graph
using the same graph-construction procedure across all acquisition
sites. The strongest connections satisfying the predefined sparsity
criterion were retained for every node. The row-wise connectivity
profile of each ROI was used as its node feature, yielding three
116-node functional graphs for every subject.

\section{Methodology}

\subsection{Problem Formulation}

Let
$\mathcal{D}_{s_1}
=
\{(\mathbf{x}^{s_1}_i,y^{s_1}_i)\}_{i=1}^{n_{s_1}}$
and
$\mathcal{D}_{s_2}
=
\{(\mathbf{x}^{s_2}_i,y^{s_2}_i)\}_{i=1}^{n_{s_2}}$
denote the two labeled source domains corresponding to Site20 and
Site21. For a selected target site, let
$\mathcal{D}_{t}
=
\{\mathbf{x}^{t}_i\}_{i=1}^{n_t}$
denote the complete target-domain dataset, whose diagnostic labels are
unavailable during training.

The objective is to learn a feature extractor $F(\cdot)$ and a
classifier $C(\cdot)$ using the two labeled source domains and the
unlabeled target domain \cite{zhao2018mdan,peng2019m3sda}. After target-domain adaptation is completed,
the concealed target labels are used only to evaluate ACC and AUC.

\subsection{Overall Framework}

\begin{figure*}[t]
\centering
\includegraphics[width=\textwidth]{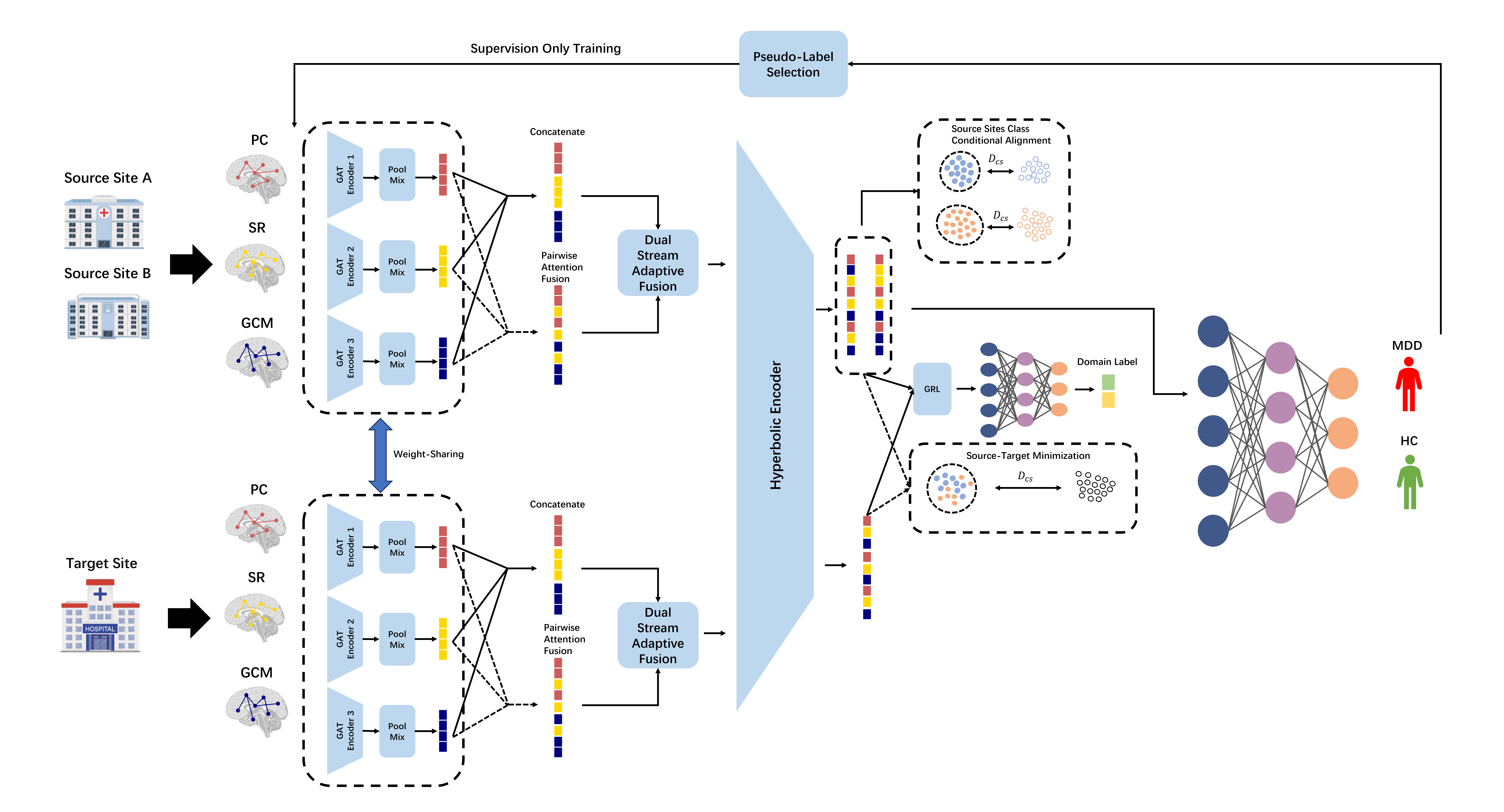}
\caption{Overview of the proposed multi-source hyperbolic multi-view
graph domain adaptation framework. PC, SR, and GCM functional graphs
are encoded by view-specific graph attention branches. Pairwise
cross-view interactions and original view representations are
integrated by dual-stream adaptive fusion, followed by a post-fusion
hyperbolic representation encoder. Source supervision and multi-source
domain adaptation are jointly used to optimize the final
representation.}
\label{fig:workflow}
\end{figure*}

Figure~\ref{fig:workflow} illustrates the proposed framework. It
contains five principal components:
\begin{enumerate}
    \item three view-specific graph attention encoders;
    \item pairwise cross-view interaction modules;
    \item a dual-stream adaptive fusion module;
    \item a post-fusion hyperbolic representation encoder; and
    \item classification and multi-source domain-adaptation objectives.
\end{enumerate}

For each subject, the PC, SR, and GCM functional graphs are first
processed by separate graph attention branches \cite{velickovic2018graph}. Pairwise interaction
modules then model dependencies between PC--SR, PC--GCM, and SR--GCM
representations. The original view representations and the cross-view
interaction representations constitute two complementary information
streams, which are integrated through sample-adaptive fusion.

The resulting Euclidean fused feature is subsequently processed by a
post-fusion hyperbolic encoder. The final representation is used
simultaneously for MDD classification, source-source alignment,
source-target adaptation, adversarial learning, information
maximization, and confidence-aware target pseudo-label supervision.

\subsection{View-Specific Graph Attention Encoding}

For each connectivity view
$v\in\{\mathrm{PC},\mathrm{SR},\mathrm{GCM}\}$, a graph attention
encoder maps the corresponding node features and weighted graph
structure to a graph-level representation:
\begin{equation}
\mathbf{h}_v
=
f_v
\left(
\mathbf{X}_v,
\mathbf{A}_v;
\boldsymbol{\theta}_v
\right),
\qquad
\mathbf{h}_v\in\mathbb{R}^{d}.
\end{equation}

Each branch contains graph attention layers, residual feature
projection, graph-level mean/max aggregation, dropout, and feature
normalization \cite{zaheer2017deepsets}. Separate branch parameters are used to preserve the
topological characteristics of the three functional connectivity
constructions.

\subsection{Pairwise Cross-View Interaction}

The three view-specific embeddings provide complementary but
heterogeneous descriptions of brain functional organization. To model
their dependencies explicitly, pairwise interaction features are
constructed as
\begin{equation}
\mathbf{q}_{uv}
=
\mathcal{I}_{uv}
\left(
\mathbf{h}_u,\mathbf{h}_v
\right),
\end{equation}
where
$(u,v)\in
\{
(\mathrm{PC},\mathrm{SR}),
(\mathrm{PC},\mathrm{GCM}),
(\mathrm{SR},\mathrm{GCM})
\}$
and $\mathcal{I}_{uv}$ denotes the corresponding learnable cross-view
attention operator.

The complete interaction stream is
\begin{equation}
\mathbf{q}
=
[
\mathbf{q}_{\mathrm{PC,SR}};
\mathbf{q}_{\mathrm{PC,GCM}};
\mathbf{q}_{\mathrm{SR,GCM}}
].
\end{equation}

In parallel, the raw-view stream preserves the original view-specific
embeddings:
\begin{equation}
\mathbf{r}
=
[
\mathbf{h}_{\mathrm{PC}};
\mathbf{h}_{\mathrm{SR}};
\mathbf{h}_{\mathrm{GCM}}
].
\end{equation}

\subsection{Dual-Stream Adaptive Fusion}

The raw-view stream and interaction stream are first projected to a
common feature space:
\begin{equation}
\tilde{\mathbf{r}}
=
\phi_r(\mathbf{r}),
\qquad
\tilde{\mathbf{q}}
=
\phi_q(\mathbf{q}).
\end{equation}

Sample-specific fusion weights are then generated by
\begin{equation}
[
\alpha_r,\alpha_q
]
=
\operatorname{softmax}
\left(
g
\left(
[
\tilde{\mathbf{r}};
\tilde{\mathbf{q}}
]
\right)
\right),
\end{equation}
where $\alpha_r+\alpha_q=1$. The fused Euclidean representation is
\begin{equation}
\mathbf{z}
=
\psi
\left(
\alpha_r\tilde{\mathbf{r}}
+
\alpha_q\tilde{\mathbf{q}}
\right).
\end{equation}

This design allows the network to adaptively balance direct
view-specific information and higher-order cross-view interactions for
each subject.

\subsection{Post-Fusion Hyperbolic Representation Encoding}

Functional brain networks exhibit modular and hierarchical
organization, which motivates the use of non-Euclidean representation
constraints. After multi-view fusion, we introduce a post-fusion
hyperbolic representation encoder based on the Poincar\'e ball model \cite{nickel2017poincare,ganea2018hyperbolic,chami2019hgcn}.
Different from applying independent geometric transformations to each
view, the encoder operates on the fused multi-view representation and
imposes a curvature-controlled constraint on the final subject-level
feature.

Let $\mathbf{z}\in\mathbb{R}^{d}$ denote the fused Euclidean feature,
and initialize $\mathbf{h}^{(0)}=\mathbf{z}$. For the $\ell$-th
hyperbolic encoding layer, a nonlinear tangent-space update is first
computed as
\begin{equation}
\mathbf{u}^{(\ell)}
=
s\cdot
\tanh
\left(
\phi_{\ell}
(
\mathbf{h}^{(\ell-1)}
)
\right),
\end{equation}
where $\phi_{\ell}(\cdot)$ is a learnable nonlinear projection with
layer normalization, ELU activation, dropout, and linear
transformation, and $s$ controls the tangent-space scale.

The tangent vector is mapped to the Poincar\'e ball with curvature
$-c$:
\begin{equation}
\tilde{\mathbf{h}}_{\mathbb{D}}^{(\ell)}
=
\operatorname{Exp}^{c}_{\mathbf{0}}
(
\mathbf{u}^{(\ell)}
).
\end{equation}

To keep the representation inside the valid Poincar\'e ball, a
curvature-dependent projection is applied:
\begin{equation}
\mathbf{h}_{\mathbb{D}}^{(\ell)}
=
\operatorname{Proj}^{c}
\left(
\tilde{\mathbf{h}}_{\mathbb{D}}^{(\ell)}
\right),
\qquad
c\|\mathbf{h}_{\mathbb{D}}^{(\ell)}\|^2 < 1 .
\end{equation}

The projected hyperbolic point is then mapped back to the tangent space
for compatibility with the classifier and alignment objectives:
\begin{equation}
\mathbf{h}^{(\ell)}
=
\operatorname{Log}^{c}_{\mathbf{0}}
\left(
\mathbf{h}_{\mathbb{D}}^{(\ell)}
\right).
\end{equation}

Finally, a learnable residual gate combines the hyperbolic-encoded
representation with the original fused feature:
\begin{equation}
\mathbf{h}
=
\operatorname{LN}
\left(
\gamma \mathbf{h}^{(L)}
+
(1-\gamma)\mathbf{z}
\right),
\end{equation}
where $\gamma$ is obtained through a sigmoid gate. The resulting feature
$\mathbf{h}$ is used for classification, Cauchy--Schwarz alignment,
adversarial adaptation, information maximization, and pseudo-label
learning.

The intermediate representation is constrained by the
curvature-dependent Poincar\'e-ball projection and subsequently read
out in the tangent space for compatibility with the downstream
objectives. Together with nonlinear tangent-space transformations and
gated residual fusion, the module provides a bounded, curvature-aware
refinement of the fused multi-view feature.

\subsection{Two-Stage Multi-Source Domain Adaptation}

\paragraph{Stage I: supervised source pretraining.}

The graph encoders, fusion module, hyperbolic encoder, and classifier
are first optimized using labeled samples from Site20 and Site21. The
source classification objective is
\begin{equation}
\mathcal{L}_{\mathrm{cls}}
=
-
\frac{1}{
n_{s_1}+n_{s_2}
}
\sum_{s\in\{s_1,s_2\}}
\sum_{i=1}^{n_s}
\sum_{c=1}^{C}
y^{s}_{ic}
\log
p^{s}_{ic}.
\end{equation}

\paragraph{Stage II: unlabeled target-domain adaptation.}

During the second stage, the complete target domain participates in
adaptation without ground-truth labels. Source classification
supervision is retained while source-source and source-target
distribution discrepancies are reduced.

\subsection{Class-Wise Cauchy--Schwarz Alignment}

Let $\mathbf{H}_{20}^{(c)}$ and $\mathbf{H}_{21}^{(c)}$ denote the final
features of source samples from class $c$ in Site20 and Site21,
respectively. Their class-conditional discrepancy is measured by
\begin{equation}
\mathcal{L}_{ss}
=
\frac{1}{C}
\sum_{c=1}^{C}
D_{\mathrm{CS}}
\left(
\mathbf{H}_{20}^{(c)},
\mathbf{H}_{21}^{(c)}
\right).
\end{equation}

The Cauchy--Schwarz divergence between two feature sets $\mathbf{X}$
and $\mathbf{Y}$ is defined as
\begin{equation}
D_{\mathrm{CS}}
(\mathbf{X},\mathbf{Y})
=
-\log
\frac{
\mathbb{E}_{\mathbf{x},\mathbf{y}}
[
k(\mathbf{x},\mathbf{y})
]
}{
\sqrt{
\mathbb{E}_{\mathbf{x},\mathbf{x}'}
[
k(\mathbf{x},\mathbf{x}')
]
\,
\mathbb{E}_{\mathbf{y},\mathbf{y}'}
[
k(\mathbf{y},\mathbf{y}')
]
}
},
\end{equation}
where $k(\cdot,\cdot)$ is an RBF kernel \cite{jenssen2006cs,gretton2012mmd}.

For confident target samples, pseudo-labels are used to construct
class-specific target feature sets $\mathbf{H}_{t}^{(c)}$. Class-wise
source-target alignment is then defined as
\begin{equation}
\mathcal{L}_{st}
=
\frac{1}{2C}
\sum_{s\in\{20,21\}}
\sum_{c=1}^{C}
D_{\mathrm{CS}}
\left(
\mathbf{H}_{s}^{(c)},
\mathbf{H}_{t}^{(c)}
\right).
\end{equation}

This formulation preserves disease-related class structure while
reducing both inter-source and source-target discrepancies.

\subsection{Domain-Adversarial Learning}

A binary domain discriminator $G_d(\cdot)$ is connected to the final
feature through a gradient reversal layer. The discriminator is trained
to distinguish source and target representations, while the feature
extractor is trained adversarially to produce domain-invariant
representations \cite{ganin2016dann,long2018cdan}. Site20 and Site21 are handled as two source domains
for source-source alignment, while the adversarial discriminator
operates on source-versus-target discrimination. The adversarial loss is
denoted by
\begin{equation}
\mathcal{L}_{\mathrm{adv}}
=
-
\sum_i
\sum_{d}
d_{id}
\log
\hat{d}_{id}.
\end{equation}

\subsection{Information Maximization}

For target prediction $\mathbf{p}_i^t$, the information-maximization
objective \cite{grandvalet2004entropy,liang2020shot} is
\begin{equation}
\mathcal{L}_{\mathrm{im}}
=
\frac{1}{n_t}
\sum_{i=1}^{n_t}
H
\left(
\mathbf{p}_i^t
\right)
-
H
\left(
\frac{1}{n_t}
\sum_{i=1}^{n_t}
\mathbf{p}_i^t
\right),
\end{equation}
where $H(\cdot)$ denotes entropy. Minimizing the first term encourages
confident predictions, while maximizing the marginal entropy prevents
all target samples from collapsing into the same class.

\subsection{Confidence-Aware Pseudo-Label Learning}

A target sample is included in pseudo-label supervision only when its
maximum predicted probability exceeds a confidence threshold $\tau$ \cite{lee2013pseudolabel,sohn2020fixmatch,liang2020shot}.
Let
\begin{equation}
\mathcal{I}_{\tau}
=
\left\{
i:
\max_c p^{t}_{ic}
\geq
\tau
\right\}.
\end{equation}

The pseudo-label loss is
\begin{equation}
\mathcal{L}_{\mathrm{pl}}
=
-
\frac{1}{
|\mathcal{I}_{\tau}|
}
\sum_{i\in\mathcal{I}_{\tau}}
\sum_{c=1}^{C}
\hat{y}^{t}_{ic}
\log
p^{t}_{ic}.
\end{equation}

The confidence threshold and the contribution of pseudo-label
supervision are progressively scheduled to reduce the influence of
incorrect target predictions during early adaptation.

\subsection{Overall Optimization Objective}

The complete adaptation objective is
\begin{equation}
\begin{aligned}
\mathcal{L}
={}&
\mathcal{L}_{\mathrm{cls}}
+
\lambda_{ss}
\mathcal{L}_{ss}
+
\lambda_{st}
\mathcal{L}_{st}
\\
&+
\lambda_{\mathrm{adv}}
\mathcal{L}_{\mathrm{adv}}
+
\lambda_{\mathrm{im}}
\mathcal{L}_{\mathrm{im}}
+
\lambda_{\mathrm{pl}}
\mathcal{L}_{\mathrm{pl}}.
\end{aligned}
\end{equation}

The loss weights are dynamically scheduled during adaptation so that
source-discriminative information is preserved while target-domain
alignment becomes progressively stronger.

\section{Experiments}

\subsection{Comparison Methods and Evaluation Metrics}

The proposed method was compared with BC-SVM, LE-SVM, DANN, UFA-Net,
MFCP, AUFA, and H2MSDA. BC-SVM and LE-SVM are conventional
functional-connectivity pipelines; DANN, UFA-Net, and AUFA are
domain-adaptation baselines \cite{long2015dan,long2018cdan}; MFCP is a multi-view non-adaptation
baseline; and H2MSDA models multiple source domains \cite{zhao2018mdan,peng2019m3sda}. For single-source
methods, Site20 only, Site21 only, and pooled Site20+Site21 settings
were evaluated. The pooled setting was used in the main comparison to
match the number of labeled samples, whereas H2MSDA and the proposed
method retained Site20 and Site21 as distinct sources. ACC and AUC were
macro-averaged across the seven target sites \cite{fawcett2006roc}.

\subsection{Implementation Details}

The model used three view-specific graph attention branches and was
optimized with AdamW \cite{loshchilov2019adamw} in two stages: source-supervised pretraining
followed by unlabeled target-domain adaptation. The pseudo-label
confidence threshold was progressively relaxed, and pseudo labels were
updated at fixed intervals. A binary discriminator aligned the pooled
source and target representations, while class-wise Cauchy--Schwarz
losses preserved inter-source and source-target class structure. The
model was implemented in PyTorch 2.6.0 \cite{paszke2019pytorch} with CUDA 11.8 and trained on an
NVIDIA GeForce RTX 4060 Laptop GPU.

\subsection{Main Cross-Site Results}

Table~\ref{tab:main_results} presents the complete target-domain
performance across the seven acquisition sites.

\begin{table*}[t]
\centering
\caption{Cross-site target-domain performance across seven sites.
Each entry reports ACC/AUC. Single-source baselines use pooled
Site20+Site21 data, whereas H2MSDA and the proposed method preserve Site20 and Site21
as distinct source domains. Bold values indicate the highest result
for each metric at each target site.}
\label{tab:main_results}
\setlength{\tabcolsep}{3.7pt}
\resizebox{\textwidth}{!}{%
\begin{tabular}{lcccccccc}
\toprule
Method
& Site6
& Site11
& Site12
& Site16
& Site19
& Site22
& Site25
& Mean\\
\midrule

UFA-Net
& 0.6000/0.6670
& 0.4920/0.6910
& 0.7110/0.4790
& 0.5320/0.4920
& 0.6670/0.6680
& 0.6000/0.6050
& 0.6180/0.6000
& 0.6028/0.6046\\

MFCP
& 0.5000/0.7470
& 0.5740/0.5660
& 0.4470/0.7710
& 0.6130/\textbf{0.6910}
& 0.5980/0.6240
& 0.6000/0.6080
& 0.4210/0.5000
& 0.5361/0.6473\\

BC-SVM
& 0.5330/0.4620
& 0.6230/0.5980
& 0.6320/0.3850
& 0.5160/0.4480
& 0.6210/0.5860
& 0.5800/0.5280
& 0.5990/0.6000
& 0.5862/0.5155\\

LE-SVM
& 0.5670/0.5200
& 0.4590/0.4940
& 0.7370/0.7080
& 0.5480/0.5860
& 0.6440/0.5820
& 0.6400/0.6310
& 0.6120/0.5000
& 0.6009/0.5832\\

DANN
& 0.5330/0.6000
& 0.5080/0.6220
& 0.7890/0.7500
& 0.5000/0.4860
& 0.6090/\textbf{0.8510}
& 0.5800/0.6920
& 0.6050/0.5000
& 0.5894/0.6472\\

H2MSDA
& 0.7000/0.6756
& 0.6393/0.6175
& 0.8421/0.7396
& 0.5806/0.6504
& 0.6897/0.7783
& 0.6800/\textbf{0.7550}
& 0.6053/0.4839
& 0.6767/0.6715\\

AUFA
& 0.5000/0.5911
& 0.5082/0.4871
& 0.8158/0.5833
& 0.5000/0.6181
& 0.5747/0.5507
& 0.6000/0.6450
& 0.5987/0.6133
& 0.5853/0.5841\\

\textbf{Ours}
& \textbf{0.7667}/\textbf{0.7511}
& \textbf{0.6885}/\textbf{0.7026}
& \textbf{0.8684}/\textbf{0.7760}
& \textbf{0.6774}/0.6597
& \textbf{0.7931}/0.8159
& \textbf{0.7200}/0.6983
& \textbf{0.6382}/\textbf{0.6294}
& \textbf{0.7360}/\textbf{0.7190}\\

\bottomrule
\end{tabular}}
\end{table*}

The proposed method achieved the highest macro-averaged ACC of
73.60\% and the highest macro-averaged AUC of 71.90\%. Compared with
H2MSDA, the strongest multi-source baseline, the proposed method
improved mean ACC by 5.93 percentage points and mean AUC by 4.75
percentage points.

The proposed method obtained the highest ACC on all seven target sites.
It also achieved the highest AUC on Site6, Site11, Site12, and Site25.
MFCP, DANN, and H2MSDA obtained the highest individual AUC on Site16,
Site19, and Site22, respectively. This site-dependent behavior
indicates that scanner- and cohort-related distribution shifts differ
substantially across target domains. Overall, the results suggest that
jointly modeling multi-view connectivity, source-domain structure, and
target-domain adaptation improves performance across the evaluated
unlabeled target sites.

\subsection{Source-Setting Sensitivity of Single-Source Methods}

To investigate the sensitivity of conventional single-source methods to
source-domain selection, each method was evaluated using Site20 only,
Site21 only, and Site20+Site21 pooled as one source domain. AUFA was
included in this analysis because aggregate results were available for
all three source configurations.

\begin{table*}[t]
\centering
\caption{Macro-averaged ACC and AUC of single-source methods under
different source-data configurations. Site20+Site21 denotes pooling the
two source datasets and treating them as a single source domain.}
\label{tab:single_source}
\setlength{\tabcolsep}{5.5pt}
\begin{tabular}{lcccccc}
\toprule
& \multicolumn{2}{c}{Site20 only}
& \multicolumn{2}{c}{Site21 only}
& \multicolumn{2}{c}{Site20+Site21 pooled}\\
\cmidrule(lr){2-3}
\cmidrule(lr){4-5}
\cmidrule(lr){6-7}
Method
& ACC & AUC
& ACC & AUC
& ACC & AUC\\
\midrule

UFA-Net
& 0.6309 & 0.4995
& 0.5487 & 0.5741
& \textbf{0.6028} & 0.6046\\

MFCP
& \textbf{0.6725} & 0.6174
& 0.5281 & 0.6231
& 0.5361 & \textbf{0.6473}\\

BC-SVM
& 0.5784 & 0.5100
& 0.5746 & 0.6043
& 0.5862 & 0.5155\\

LE-SVM
& 0.5930 & 0.4669
& 0.6060 & 0.6287
& 0.6009 & 0.5832\\

DANN
& 0.5902 & \textbf{0.6236}
& 0.5075 & \textbf{0.6616}
& 0.5894 & 0.6472\\

AUFA
& 0.5499 & 0.5366
& \textbf{0.6246} & 0.6421
& 0.5853 & 0.5841\\

\bottomrule
\end{tabular}
\end{table*}

The results indicate that single-source methods are highly sensitive to
source-domain selection. No method consistently achieved the best
performance under all three configurations. Moreover, directly pooling
Site20 and Site21 did not uniformly improve performance. This finding
suggests that simply increasing the number of labeled source samples
does not adequately address inter-source distribution discrepancy.

\subsection{Single-Source and Dual-Source Analysis}

The proposed architecture was further evaluated using Site20 only,
Site21 only, and the complete Site20+Site21 dual-source configuration.
For the single-source variants, the same multi-view and hyperbolic
architecture was retained, while the second source domain and
inter-source alignment were removed.

\begin{table}[t]
\centering
\caption{Source-domain configuration analysis of the proposed
architecture.}
\label{tab:source_config}
\begin{tabular}{lcc}
\toprule
Source setting & Mean ACC & Mean AUC\\
\midrule
Site20 only
& 0.7012
& 0.6627\\

Site21 only
& 0.6751
& 0.6017\\

Site20+Site21
& \textbf{0.7360}
& \textbf{0.7190}\\
\bottomrule
\end{tabular}
\end{table}

The dual-source configuration improved ACC and AUC by 3.48 and 5.63
percentage points over Site20 only, and by 6.09 and 11.73 points over
Site21 only. These results support the complementarity of the two
source domains and the benefit of preserving their domain identities.

\subsection{Ablation Studies}

Table~\ref{tab:ablation} evaluates the contributions of domain
adaptation, post-fusion hyperbolic representation encoding, and the
three connectivity views.

In the naming of the single-view variants, ``DA'' denotes the complete
domain-adaptation objective and ``Hyp'' denotes the hyperbolic encoder.
For example, PC+DA+Hyp retains only the PC connectivity view while
preserving both domain adaptation and hyperbolic representation
encoding.

\begin{table}[t]
\centering
\caption{Macro-averaged component and single-view ablation results.}
\label{tab:ablation}
\small   
\begin{tabular}{lcc}
\toprule
Variant & Mean ACC & Mean AUC\\
\midrule
Full model
& \textbf{0.7360}
& \textbf{0.7190}\\

No domain adaptation
& 0.6561
& 0.5258\\

w/o hyperbolic encoder
& 0.6486
& 0.5456\\

\midrule
PC + DA + Hyp
& 0.6829
& 0.6372\\

PC + Hyp
& 0.6634
& 0.5881\\

PC only
& 0.6638
& 0.5810\\

\midrule
SR + DA + Hyp
& 0.6606
& 0.5862\\

SR + Hyp
& 0.7034
& 0.6696\\

SR only
& 0.6480
& 0.5538\\

\midrule
GCM + DA + Hyp
& 0.6407
& 0.5165\\

GCM + Hyp
& 0.6347
& 0.4974\\

GCM only
& 0.6290
& 0.5085\\

\bottomrule
\end{tabular}
\end{table}

Removing domain adaptation reduced mean ACC/AUC from 0.7360/0.7190 to
0.6561/0.5258, while removing the hyperbolic encoder reduced them to
0.6486/0.5456. The strongest single-view variant was SR+Hyp, achieving
0.7034 ACC and 0.6696 AUC, but it remained below the complete model by
3.26 and 4.94 percentage points, respectively. Overall, the full
three-view configuration achieved the strongest aggregate performance
among the evaluated variants.

\subsection{Representation Visualization}

\begin{figure}[t]
\centering
\includegraphics[
width=0.8\textwidth
]{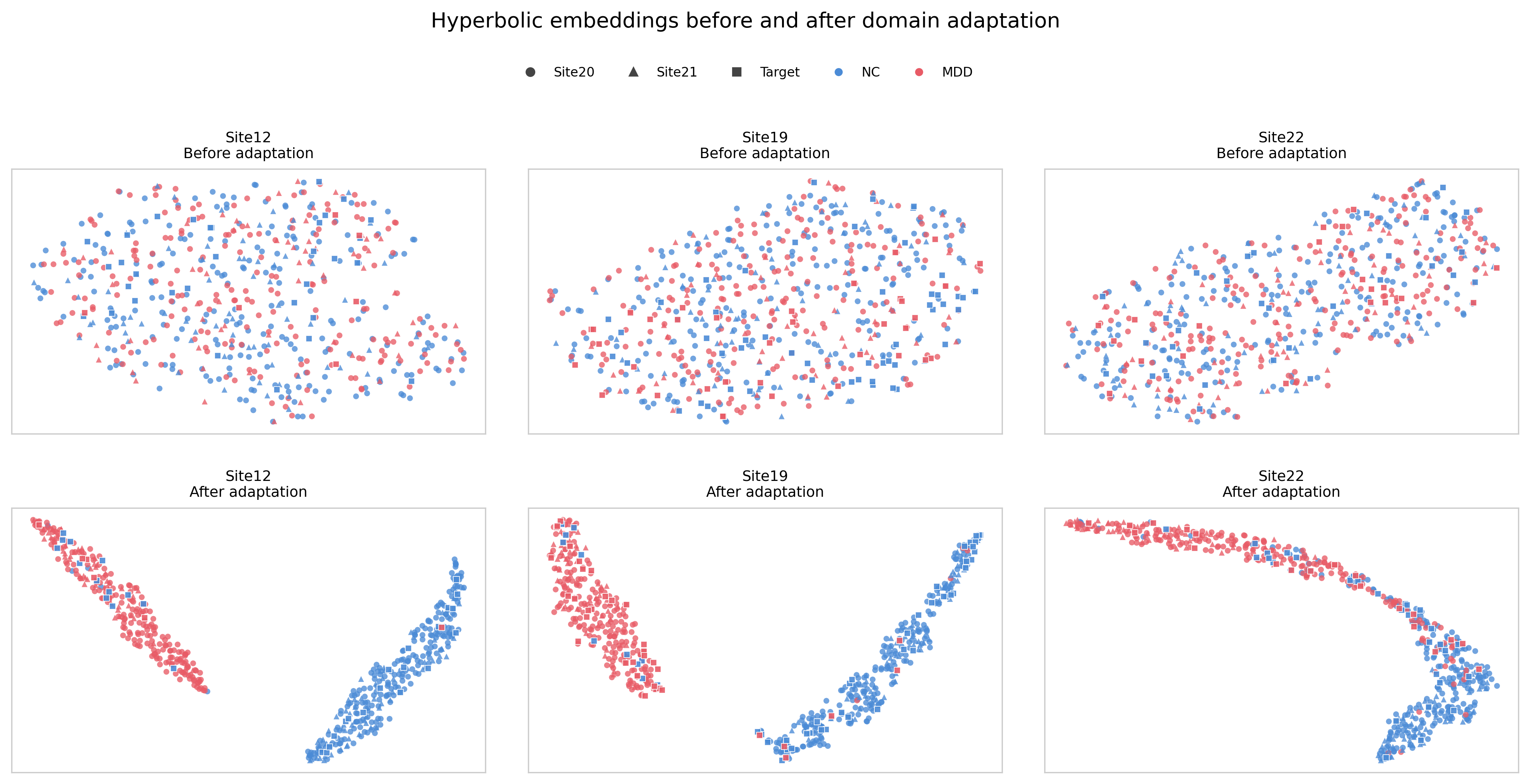}
\caption{Joint t-SNE visualization of post-fusion hyperbolic
representations before and after domain adaptation for Site12, Site19,
and Site22. Colors denote diagnostic classes (blue: HC; red: MDD),
while marker shapes denote domains (circle: Site20; triangle: Site21;
square: target site). After adaptation, samples from different domains
are more consistently organized within class-related manifolds, and
the separation between HC and MDD becomes more pronounced.}
\label{fig:tsne}
\end{figure}

Figure~\ref{fig:tsne} \cite{vandermaaten2008tsne} compares the post-fusion representations before
and after target-domain adaptation.

Before adaptation, the embeddings exhibited substantial class overlap
and site-dependent variation. After adaptation, they showed clearer
class-related organization across the source and target domains.
Because t-SNE is nonlinear, the visualization is treated as qualitative
supporting evidence rather than a quantitative measure of alignment.

\section{Conclusion}

We introduced a unified multi-source, multi-view graph domain adaptation
framework for cross-site MDD identification from rs-fMRI. The method
preserves view-specific graph structures, explicitly models pairwise
cross-view dependencies, adaptively fuses direct and interaction-based
representations, and refines the fused feature through lightweight
hyperbolic residual encoding. A two-stage transductive adaptation strategy
combines class-conditional inter-source and source-target alignment with
adversarial learning, information maximization, and confidence-aware
pseudo-labeling.

Across seven independently adapted target sites, the framework achieved
macro-averaged ACC/AUC of 73.60\%/71.90\%, exceeding the strongest evaluated
baseline by 5.93/4.75 percentage points. Source-configuration and ablation
analyses further support the value of preserving source identities,
integrating heterogeneous FC views, and applying post-fusion geometric
refinement. The current evidence is limited to the evaluated transductive
protocol; future work will examine independent-site validation, uncertainty
estimation, multimodal integration, and clinically grounded interpretation.

\bibliography{aaai2027}


\end{document}